\documentclass[a4paper,fleqn]{cas-dc}

\usepackage[authoryear,longnamesfirst]{natbib}
\newcommand{\etal}{\textit{et al.}}
\def\tsc#1{\csdef{#1}{\textsc{\lowercase{#1}}\xspace}}
\tsc{WGM}
\tsc{QE}
\tsc{EP}
\tsc{PMS}
\tsc{BEC}
\tsc{DE}

\begin{document}
\let\WriteBookmarks\relax
\def\floatpagepagefraction{1}
\def\textpagefraction{.001}

\shorttitle{Human Health Indicator Prediction from Gait Video}

\shortauthors{Ziqing Li et~al.}

\title [mode = title]{Human Health Indicator Prediction from Gait Video}          

\author[1]{Ziqing Li}\cormark[2] 
\address[1]{Tsinghua University, Haidian District, Beijing 100084, China}
\author[1]{Xuexin Yu}
\author[1]{Xiaocong Lian}
\author[2]{Yifeng Wang}
\address[2]{HIT Campus of University Town of Shenzhen, Shenzhen 518055, China}
\author[1]{Xiangyang Ji}
\cormark[1] 
\cortext[cor1]{Corresponding author}
\tnotetext[2]{lizq21@mails.tsinghua.edu.cn (Z.Li)}

\begin{abstract}
Body Mass Index (BMI), age, height and weight are important indicators of human health conditions, which can provide useful information for plenty of practical purposes, such as health care, monitoring and re-identification. Most existing methods of health indicator prediction mainly use front-view body or face images. These inputs are hard to be obtained in daily life and often lead to the lack of robustness for the models, considering their strict requirements on view and pose. In this paper, we propose to employ gait videos to predict health indicators, which are more prevalent in surveillance and home monitoring scenarios. However, the study of health indicator prediction from gait videos using deep learning was hindered due to the small amount of open-sourced data. To address this issue, we analyse the similarity and relationship between pose estimation and health indicator prediction tasks, and then propose a paradigm enabling deep learning for small health indicator datasets by pre-training on the pose estimation task. Furthermore, to better suit the health indicator prediction task, we bring forward \textbf{G}lobal-\textbf{L}ocal \textbf{A}ware a\textbf{N}d \textbf{C}entrosymmetric \textbf{E}ncoder (GLANCE) module. It first extracts local and global features by progressive convolutions and then fuses multi-level features by a centrosymmetric double-path hourglass structure in two different ways.

Experiments demonstrate that the proposed paradigm achieves state-of-the-art results for predicting health indicators on MoVi, and that the GLANCE module is also beneficial for pose estimation on 3DPW.
\end{abstract}



\maketitle

\section{Introduction}
\label{sec:intro}

The Body Mass Index (BMI), height, weight and age are important indicators of one's health conditions, where BMI is defined as
\begin{equation}
  BMI=\frac{weight[kg]}{{height}^2[m^2]}=\frac{weight[lb]\times 703}{{height}^2[{in}^2]}. 
\end{equation}
For example, BMI, as an integrated variable of weight and height, is found to be related to cancers, unstable angina, myocardial infarction, type \uppercase\expandafter{\romannumeral2} diabetes and cardiovascular disease \cite{renehan2008body, wolk2003body, meigs2006body}. Moreover, age is a widely recognized indicator for health status evaluation and almost all disease diagnoses. Therefore, accurately predicting BMI, height, weight and age in a contactless manner can be helpful for daily monitoring of health conditions and primary screening for diseases. In addition, these health indicators are beneficial for the police department as they are frequently used in surveillance, forensics and re-identification applications\cite{bmiforid}.

Most existing works on health indicator prediction are conducted on the frontal-view images of body or face using manual feature extraction and deep learning methods \cite{imagebmi1, PASCALI2016238, WEN2013392,2017Face,2021Estimation}. Traditional methods using manual feature extraction emphasizes on domain knowledge, constraining the upper limit of the algorithm performance to human capability \cite{imagebmi1, PASCALI2016238, WEN2013392}. Meanwhile, deep learning methods pushes boundaries by letting the network learn features itself. However, these methods are sensitive to viewpoint change and some require complex pre-processing steps \cite{jin2022attention}. 

In this work, we introduce an end-to-end deep learning algorithm to predict health indicators from gait videos. Compared with existing approaches, our scheme has three advantages. First, it directly extracts related features from captured gait videos in an end-to-end manner, without needing prior knowledge. Second, gait videos are more convenient and easy to be obtained in real-life scenarios, such as surveillance and short videos. Predicting health status from gait videos is promising for home health monitoring in the Internet of Things era. Third, the methods based on gait videos are less sensitive to view changes than those based on frontal body or facial images. However, the datasets containing human gait with health status indicators are either relatively small or not available to the public, which would make it difficult to directly train a well-functioning deep learning network. Fortunately, monocular-view video-based pose estimation can provide useful information for health indicator prediction \cite{rossoInfluenceBMIGait2019,shinEffectMuscleStrength2014,vanderstraatenMobileAssessmentLower2018,windhamImportanceMidtoLateLifeBody2017}.  It is also a rather developed field of study with many large-scale and publicly-available datasets \cite{human36,Muco,surreal}. Based on the relationship of both tasks, transfer learning is introduced to the health status prediction task to overcome the issue of insufficient data. The backbone of the model is first pre-trained on pose estimation datasets and then transferred into the health status prediction task.

Our main contributions are summarized as follows:
\begin{itemize}
    \item To the best of our knowledge, we make the first attempt to use deep learning on gait videos to predict health indicators.
    \item We present a paradigm enabling deep learning for a small health indicator dataset by pre-training on the pose estimation task. 
    \item We propose a global-local aware and centrosymmetric encoder (GLANCE) to extract spatial features, which focuses on the extraction and integration of multi-level features.
    \item Experiments demonstrate that the proposed method achieves state-of-the-art results for predicting health indicators on MoVi \cite{ghorbani2021movi}, and that the GLANCE module is also beneficial for pose estimation on 3DPW.
\end{itemize}

The rest of this paper is organized as follows. Section \ref{related work} summarizes related works, including health indicators prediction, the relationship between pose estimation and health indicator prediction, and transfer learning. Section \ref{methodology} first elaborates the proposed global-local aware and centrosymmetric encoder (GLANCE) module, and then describes the architecture and pipeline of the proposed GlanceNet. Section \ref{experiments} discusses implementing details briefly and demonstrates the superior performance of the proposed GlanceNet compared with the state-of-the-art method on the MoVi. Furthermore, the ablation study is investigated in this section. Finally, the summary of the paper is presented in Section \ref{summary}.

\section{Related work}\label{related work}

Currently, most methods estimate health indicators from facial and body images, which can be roughly categorized into conventional and deep learning methods according to their way of extracting features.
Conventional methods extract features in a computational and manual way relying on prior domain knowledge \cite{imagebmi1, PASCALI2016238, WEN2013392}. For example, Wen and Guo \cite{WEN2013392} detect keypoints in frontal-view face image and use their coordinates to calculate pre-defined features, cheekbone to jaw width (CJWR), width tupper facial height ratio (WHR), perimeter to area ratio (PAR) etc. These computational features serve as input to the support vector regression for BMI prediction. 
Whereas deep learning methods automatically learns how to extract features. With the rapid development of deep learning techniques, these deep learning methods have outperformed the conventional ones in both the face image and front body image prediction tasks.
Kocabey \etal \cite{2017Face} use a pre-trained backbone network to extract features from face images. Improving on their work, Yousaf \etal \cite{2021Estimation} proposed Region aware Global Average Pooling (Reg-GAP), which pools the feature maps from pre-trained backbone networks according to their corresponding face regions, eye, nose, eyebrow, lips, etc. 

From the perspective of ergonomics and medicine, various studies have validated that gait pose could reflect human inner health status\cite{rossoInfluenceBMIGait2019, shinEffectMuscleStrength2014, vanderstraatenMobileAssessmentLower2018, windhamImportanceMidtoLateLifeBody2017}. For instance, Zhong \etal 
 \cite{zhongApplicationSmartBracelet2018} calculated several features of gait with wearable sensors and found that pre-frail older adults showed  a decrease in speed and increases in RMS and step irregularity significantly compared with the non-frail counterparts. Calvache \etal  \cite{fall1} have successfully utilized pose estimation methods to predict the balance and physical equilibrium of the human body from videos, in order to prevent falls. Moccia \etal  \cite{infant} uses one infant's limb joint information from pose estimation to assess its cognitive development. Therefore, for health status evaluation, it is worth looking into the research of pose estimation.

 Among the pose estimation methods, the one most relative to health indicator prediction objectives is the monocular-view video-based pose estimation. One of the most influential works in this domain is the introduction of a parametric model Skinned Multi-person Linear Model (SMPL) \cite {2015SMPL}. The information on body shape and pose variation are summarized and reduced to pose, shape and camera parameters in the SMPL model. Using this parametric model, one can easily reconstruct a realistic human body mesh. Plenty of pose estimation studies are carried out based on the SMPL model.For example, VIBE \cite{kocabas_vibe_2020} is an important benchmark, which is the first one to utilize adversarial learning to incorporate prior knowledge into pose estimation from video. Some works do not rely on parametric model and try to regress mesh vertices and joints coordinates directly from images \cite{openpose,hrnet,Metro}. METRO \cite{Metro} uses transformer to attend to the interactions between joints and vertices, in order to accurately reconstruct human body from an image. Compared with health status prediction from gait videos, pose estimation is a relatively developed area of research and provides many large-scale and publicly-available datasets \cite{human36,Muco,surreal}. This makes it possible to train deep neural networks firstly on pose estimation datasets and then transfer learned weights into health status prediction models.

Transfer learning can take advantage of similarities between tasks, such as applying a model trained on bicycles to motorcycles, or a model trained on cats to dogs. Since there is little difference in features between bikes and motorcycles, cats and dogs, transfer learning can often achieve good results, especially in the absence of a certain number of data, and the use of sufficient similar data can largely compensate for this deficiency. In today's research, the weights pre-trained on ImageNet \cite{Imagenet} have been widely transferred to many fields, e.g., Transferring GANs \cite{Transfer_GAN} achieve good image style transfer based on limited data using the weights pre-trained on ImageNet. DWGAN \cite{DWGAN} achieves favorable results on the image deblurring problem and overcomes the problem of insufficient data. Inspired by these tasks and previous studies on the correlation between human posture estimation and health status, we propose to transfer the weights of human pose estimation to human health status prediction.

\section{Methodology} \label{methodology}

\begin{figure*}[h]
\begin{center}
\includegraphics[width=0.98\textwidth]{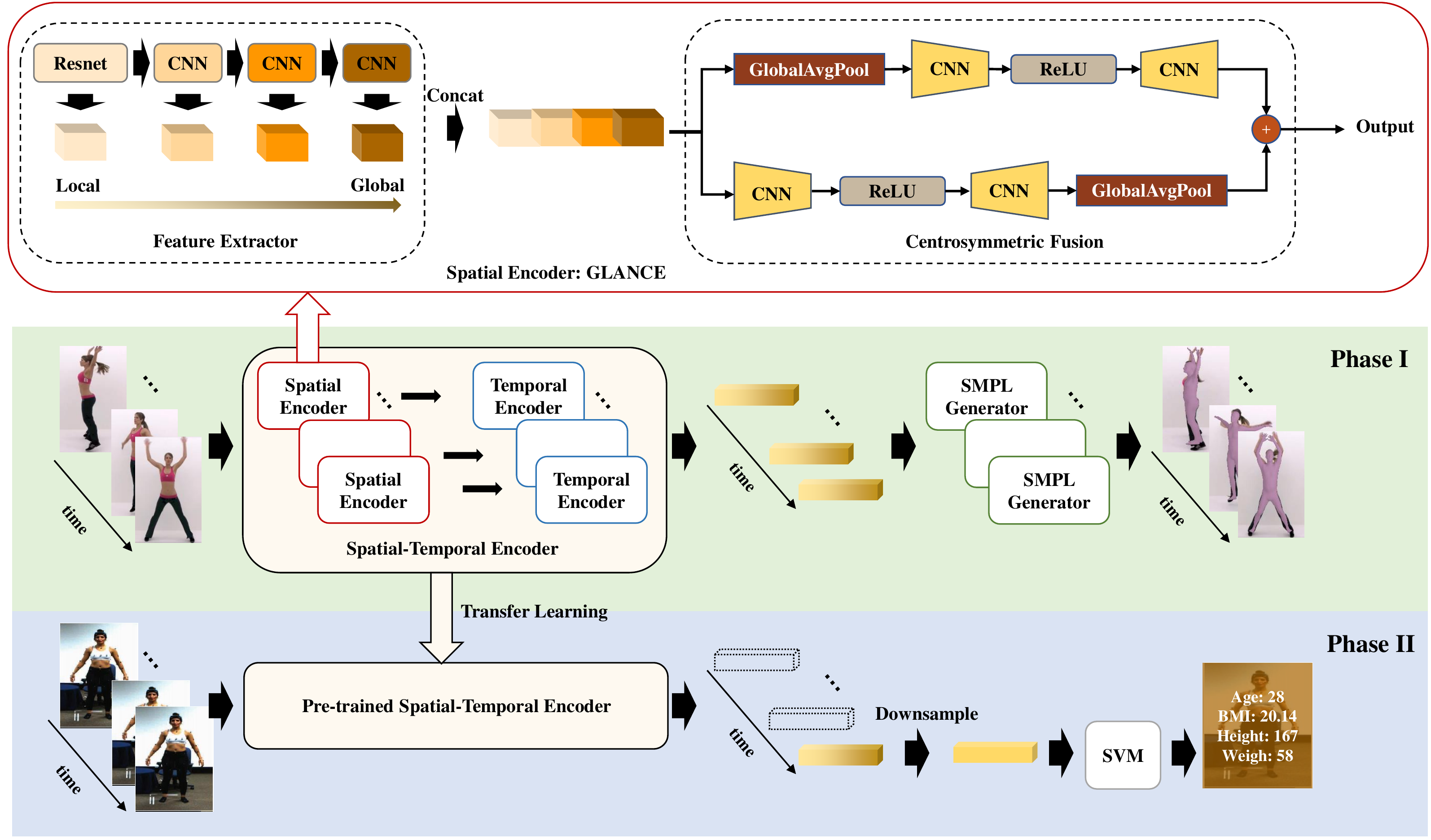}
\end{center}
  \caption{The architecture and pipeline of the proposed GlanceNet, which consists of two phases. In phase \uppercase\expandafter{\romannumeral1}, the model is trained on a large pose estimation dataset to generate SMPL parameters for every frame in the input video, using a spatial-temporal encoder and a SMPL generator. In phase \uppercase\expandafter{\romannumeral2}, the spatial-temporal encoder trained in phase \uppercase\expandafter{\romannumeral1} is employed to extract features for the small MoVi dataset. The features are then downsampled and fed into a SVM to predict health indicators. The spatial encoder GLANCE in the spatial-temporal encoder is designed to extract and incorporate local and global information.}
\label{fig:pipeline}
\end{figure*}

Motivated by insufficient datasets for human gait with health status indicators, we introduce transfer learning to the health status prediction task. Therefore, the proposed GlanceNet, shown in Figure \ref{fig:pipeline}, is divided into two phases. In phase \uppercase\expandafter{\romannumeral1}, we perform training on a large dataset for the human pose estimation task. In phase \uppercase\expandafter{\romannumeral2}, the well-trained encoder in the previous phase serves as a feature extractor for the tiny health indicator prediction dataset. To take full advantage of the pose estimation task, the network architecture of phase \uppercase\expandafter{\romannumeral1} follows the VIBE \cite{kocabas_vibe_2020} model, which mainly consists of a spatial-temporal encoder (a ResNet and a GRU unit) and a SMPL Generator. Here, we choose the SMPL model as optimization objective for the pose estimation task. The main reason is that the parameters of the SMPL model include information about body shape and joint locations, which is directly related to health indicators like height and weight. Considering characters of videos in space and time, a spatial-temporal encoder is indispensable to extract features from intra- and inter-frame. In addition, to improve the prediction of health indicators, we propose a spatial encoder, \textbf{G}lobal-\textbf{L}ocal \textbf{A}ware a\textbf{N}d \textbf{C}entrosymmetric \textbf{E}ncoder (GLANCE), to extract and integrate local and global spatial features. The spatial and temporal encoders are further discussed in Section \ref{sec:21} and \ref{sec:temperal}, respectively. Then the overall pipeline of our GlanceNet is explained in detail in Section \ref{sec:22}.

\subsection{Spatial Encoder: GLANCE}
\label{sec:21}
For health indicator prediction, it is of great importance to extract feature representation with local- and global-awareness from each frame. For example, height is the length between the top of the head and the feet, which is calculated across global human body features. The estimation of weight needs to obtain the joint position and body information, which are closely related to both local and global information. To meet the above requirements, we elaborately design a spatial encoder, GLANCE. As shown at the top of Figure \ref{fig:pipeline}, the GLANCE module can be further divided into feature extractor and centrosymmetric fusion components, which bear the responsibilities of extracting global-local features and fusing multi-level features, respectively. 

As shown in Figure \ref{fig:pipeline}, the feature extractor is composed of ResNet and three convolutions, which is inspired by Kim \etal \cite{fpn} and Artacho \etal \cite{artacho2020unipose}. Here, ResNet is used to extract local features and global features are extracted progressively by the following three convolutions, and all the feature maps from every convolutional neural network (CNN) layer are concatenated to form a representation with both global and local information. In addition, we also use dilated convolutions to quickly enlarge the receptive field without introducing more CNN layers. In fact, compared to algorithms like transformers, three convolutions following the ResNet extract features to be relative local scale. Nevertheless, convolutions are comparably more lightweight and requires fewer computation resources, which is advantageous considering the size of input videos. 

To fuse the stacked features from the feature extractor, we propose a centrosymmetric fusion, as shown in Figure \ref{fig:pipeline}. In centrosymmetric fusion, the two basic building blocks are an hourglass channel-wise architecture and a global average pooling with depth-wise convolution, where the former is responsible for channel-wise feature fusion and the latter accomplishes feature integration in space. Concretely, in the top path, since features first go into global average pooling with depth-wise convolution and then hourglass network, the features are fused at the global level due to losing most local information in global average pooling process. Therefore, the obtained features enhance the global information yet may ignore local information. To avoid the drawback, the bottom path is designed. When the features first walk through the hourglass network, some local features may be emphasized since current features contain all local information. The output features also highlight some local information after global average pooling with depth-wise convolution. The two paths in centrosymmetric fusion complement each other and explore the incorporation of global and local information.

\subsection{Temporal Encoder}
\label{sec:temperal}
Temporal information among frames is significant for health indicator prediction tasks since it provides information associated with motion velocity and acceleration, which is closely related to some heath indicators like age. Moreover, when the viewpoint changes and certain limbs maybe occluded in some frames of gait videos, temporal information is beneficial for making more informed judgements by comprehending the whole video sequence.
Inspired by the success of the Gated Reccurent Unit (GRU) in machine translation task by Cho \etal \cite{GRU}, we use bi-directional GRU as the temporal encoder, which makes the feature vectors corresponding to different timestamp learn from each other and the model more robust to viewpoint changes and occlusion.

\subsection{Overall Pipeline: GlanceNet}
\label{sec:22}
In this section, we will expand on the description of the pipeline GlanceNet in Figure \ref{fig:pipeline} and walk through the phase \uppercase\expandafter{\romannumeral1} and phase \uppercase\expandafter{\romannumeral2} design in detail.

\begin{figure}
	\centering
		\includegraphics[scale=.4]{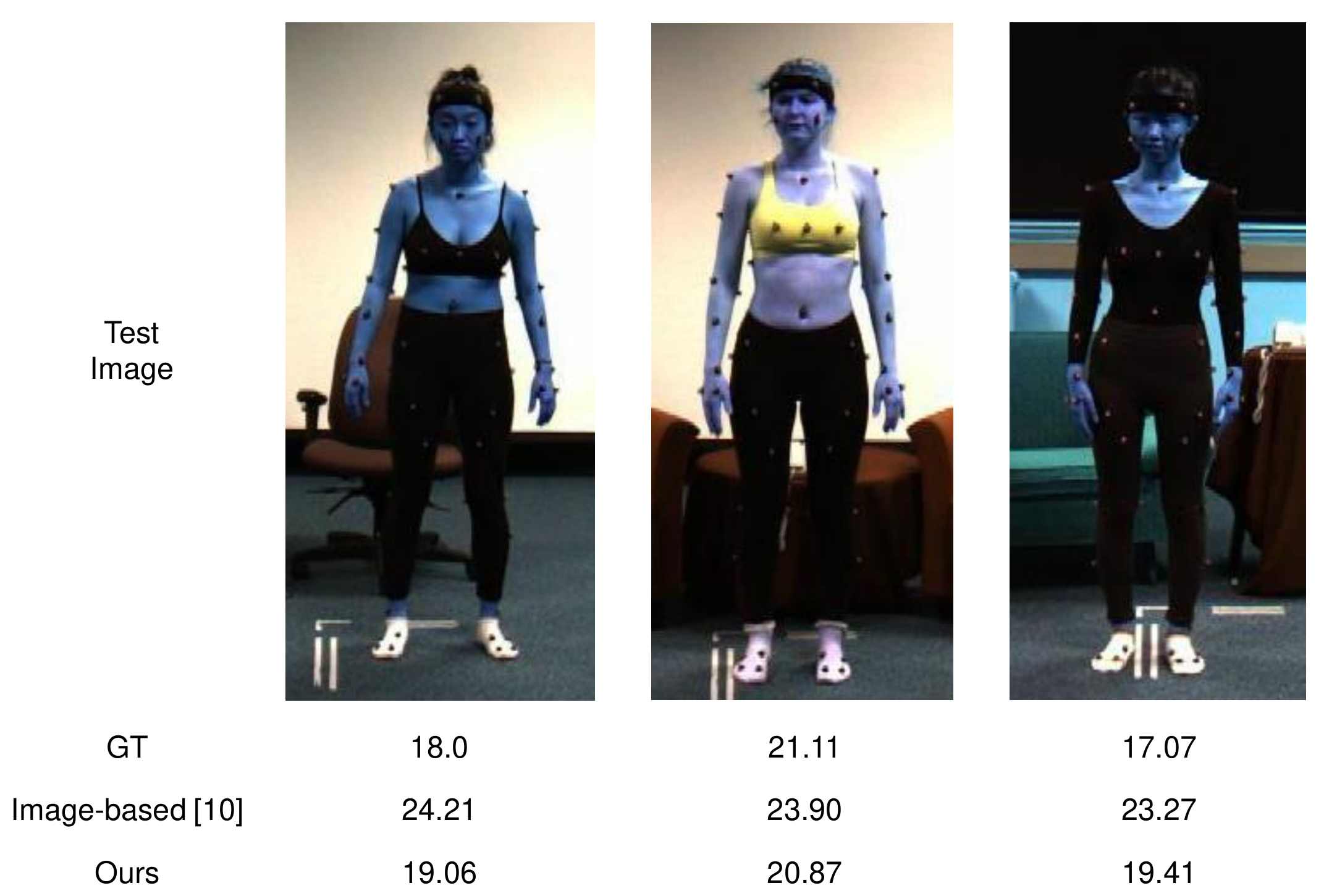}
	\caption{Examples of different methods of estimating BMI}
	\label{black}
\end{figure}

\begin{table*}[width=.9\textwidth,cols=4,pos=h]
  \caption{Quantitative comparison for health indicators prediction.}
  \begin{tabular*}{\tblwidth}{@{} LLLLLL@{} }
  \toprule
    \multirow{2}*{Models} &\multirow{2}*{ Metrics} & \multicolumn{4}{c}{Health Indicators}\\
    \cline{3-6}
         & & BMI & Age & Height & Weight \\
  \midrule
    \multirow{2}*{Image-based model \cite{jin2022attention}} & MAE $\downarrow$ & 2.61 & - & - & -\\
        \cline{2-6}
        & MAPE $\downarrow$ & 11.45\% & - & - & -\\
        \hline
        \multirow{2}*{Our Model} & MAE $\downarrow$ & 2.29 & 2.09 & 5.16 & 6.96\\
        \cline{2-6} 
        & MAPE $\downarrow$ & 9.86\% & 9.20\% & 3.09\% & 10.72\%\\
  \bottomrule
  \end{tabular*}
  \label{tab:phaseII}
\end{table*}

In phase \uppercase\expandafter{\romannumeral1}, we train the network on the pose estimation task with large publicly-available datasets. We directly feed every input video frame into the spatial encoder GLANCE, which outputs a feature vector composing of spatial information. Then the spatial feature vectors of all frames are ordered to a time sequence and fed into the temporal encoder GRU for temporal information learning, producing a stream of feature vectors of both spatial and temporal information. Finally, the feature vector of every frame is fed into the SMPL generator to produce SMPL parameters for every frame. The SMPL generator and the loss function follow the same design as VIBE \cite{kocabas_vibe_2020}. The SMPL generator contains a pre-trained regressor and a generative adversarial network to incorporate prior knowledge about the human body in AMASS dataset. 

In phase \uppercase\expandafter{\romannumeral2}, we will predict health indicators from gait videos. The dataset for phase \uppercase\expandafter{\romannumeral2} is minimal in size. Therefore, considering the similarity between pose estimation and health indicator prediction tasks stated in Section \ref{sec:intro}, the pre-trained spatial-temporal encoder trained in phase \uppercase\expandafter{\romannumeral1} is used to extract features for the health indicator prediction task. For the output of the spatial-temporal encoder, only the feature vector of the last frame is utilized since it is expected to contain information of the whole video sequence considering the GRU design and the health indicator prediction task predicts one label for the entire video sequence instead of for every frame.
Also, due to the small dataset of the health indicator prediction task, we first downsample the original features using average pooling operation, and then feed it into Support Vector Machine (SVM) regressor, which is good at handling datasets with small sample yet high-dimensional input.

\section{Experiments and Results}\label{experiments}
In this section, We first describe the experimental setup. Subsequently, we compare our method with state-of-the-art methods on MoVi dataset. Finally, ablation studies are conducted to verify the effectiveness of the proposed GLANCE module.

\subsection{Experimental Setup}
For the pose estimation task in phase \uppercase\expandafter{\romannumeral1}, three pose estimation datasets PennAction \cite{zhou2019penaction}, PoseTrack \cite{andriluka2018posetrack} and 3DPW \cite{von20183dpw} are used for training and evaluation is performed on 3DPW. The three datasets contain video recordings of people doing daily activities. PennAction and PoseTrack have 2D ground truth keypoint annotations. 3DPW dataset has 3D ground truth keypoint labels as well as annotation of SMPL parameters.
In phase \uppercase\expandafter{\romannumeral2}, the health indicator prediction task is trained and tested on MoVi dataset \cite{ghorbani2021movi}. MoVi contains gait video sequences and there are a total of 87 people with available health indicator annotations. These health indicators include age, weight, height and BMI, whose statistical information is shown in Table \ref{tab:statis}. The strategy of 5-fold cross-validation is used, and training and testing set ratios are 4:1.

The proposed model is implemented in PyTorch. The model in phase \uppercase\expandafter{\romannumeral1} is trained on eight NVIDIA GeForce RTX3090 GPUs with a batch size of 24. Its weights are initialized and optimized by the kaiming method \cite{he2015delving} and the Adam algorithm \cite{kingma2014adam} with $\beta_1$ = 0.900, respectively. During 30 epochs, the initial learning rate is $1e-3$, then is divided by 10 times after 5 epochs. In phase \uppercase\expandafter{\romannumeral2}, the SVM regressor is only trained on the 3DPW dataset as the parameters of spatial-temporal encoder are transferred from phase \uppercase\expandafter{\romannumeral1}.
\begin{table*}[width=.9\textwidth,cols=8,pos=h]
  \caption{Ablation study for health indicator prediction.}
  \begin{tabular*}{\tblwidth}{@{} LLLLLLLL@{} }
  \toprule
    \multicolumn{3}{c}{Modules} & \multirow{2}*{Metrics} & \multicolumn{4}{c}{Health Indicators}\\
        \cline{1-3} 
        \cline{5-8}
        ResNet & Extractor & Fusion &  & BMI & Age & Height & Weight\\
  \midrule
    \multirow{2}*{\checkmark}& \multirow{2}*{} & \multirow{2}*{} & MAE $\downarrow$ & 2.6   & 3.0   & 7.5   & 8.9  \\
      \cline{4-8}
        & & & MAPE $\downarrow$ & 11.2\% & 13.4\% & 4.5\% & 13.9\%\\
        \hline
        \multirow{2}*{\checkmark}& \multirow{2}*{\checkmark} & \multirow{2}*{} & MAE $\downarrow$ & 2.5   & 2.3   & 5.8   & 7.7  \\
      \cline{4-8}
        & & & MAPE $\downarrow$ & 10.7\% & 10.3\% & 3.5\% & 11.8\%\\
        \hline
        \multirow{2}*{\checkmark}& \multirow{2}*{\checkmark} & \multirow{2}*{\checkmark} & MAE $\downarrow$ & 2.3   & 2.1   & 5.2   & 7.0  \\
      \cline{4-8}
        & & & MAPE $\downarrow$ & 9.9\% & 9.2\% & 3.1\% & 10.7\%\\
  \bottomrule
  \end{tabular*}
  \label{ablation_health}
\end{table*}
        

\begin{table*}[width=.9\textwidth,cols=7,pos=h]
  \caption{Ablation study for pose estimation. The model with only the Resnet module as encoder is equivalent to VIBE\cite{kocabas_vibe_2020}.}
  \begin{tabular*}{\tblwidth}{@{} LLLLLLL@{} }
  \toprule
    \multicolumn{3}{c}{Modules} & \multicolumn{4}{c}{Metrics}\\
        \cline{1-3} 
        \cline{4-7}
        ResNet & Extractor & Fusion & MPJPE $\downarrow $  & PA-MPJPE $\downarrow$  & PVE $\downarrow$  & LimbLen Error $\downarrow$ \\
  \midrule
    \checkmark & & & 111.4 & 70.3 & 129.6& 333.5\\
        \hline
        \checkmark & \checkmark & & 109.5 & 69.7 &131.4 & 299.1\\
        \hline
        \checkmark & \checkmark &\checkmark & 105.9 & 67.2 & 127.2  & 292.8 \\
  \bottomrule
  \end{tabular*}
  \label{ablation_pose}
\end{table*}


\begin{table}
  \caption{Health indicator statistics.}
  \begin{tabular*}{\tblwidth}{@{} LLLL@{} }
  \toprule
   Health Indicators & Average & Standard Deviation \\
  \midrule
    Age & 21.75 & 3.77 \\
    Height & 168.84 & 8.93 \\
    Weight & 64.87 & 11.07 \\
    BMI & 22.71 & 3.21 \\
  \bottomrule
  \end{tabular*}
  \label{tab:statis}
\end{table}

For health indicator prediction, phase \uppercase\expandafter{\romannumeral2}, the models are evaluated by the Mean Absolute Percentage Error (MAPE) and Mean Absolute Error (MAE) of each health indicator. To demonstrate the effectiveness of the proposed GLANCE module in pose estimation task, common pose estimation metrics, Procrustes-Aligned Mean Per Joint Position Error (PA-MPJPE), Mean Per Joint Position Error (MPJPE) and Per Vertex Error (PVE), are used to evaluate the performance of models in phase \uppercase\expandafter{\romannumeral1}. In addition, we introduce a new metric, LimbLen Error, to reflect the relationship between pose estimation and health indicator prediction. Specially, the metric calculates the error of the total length of limbs between prediction and ground truth, where the length of a limb is defined as the distance between adjacent joint locations in 3D coordinates. Therefore, LimbLen is a factor to bridge through both tasks. 




\subsection{Comparisons with state-of-the-art methods}
Since our method is the first to use gait videos to predict relevant health indicators, there are no other video-based approaches to be compared against it. Currently, as discussed in Section \ref{sec:intro}, most related works mainly use single-frame image to predict health status. One of the most recent methods is the attention guided end-to-end BMI estimation network \cite{jin2022attention}, which estimates the health indicators from a frontal-view instance image. To be a fair comparison, we first select front-view images of subjects in stance position from the videos of the MoVi dataset according to the requirements \cite{jin2022attention}. Then these selected images are pre-processed to remove the background and segment human figures before feeding into the pre-trained model released officially. Table \ref{tab:phaseII} shows the corresponding quantitative comparison between both methods, where the image-based model refers to the guided end-to-end BMI estimation network \cite{jin2022attention}. Our method outperforms the image-based model \cite{jin2022attention} in MAE and MAPE for the BMI indicator, which indicates that our method extracts more effective features from videos than the image-based model. 

\begin{figure*}
	\centering
	  \includegraphics[scale=0.4]{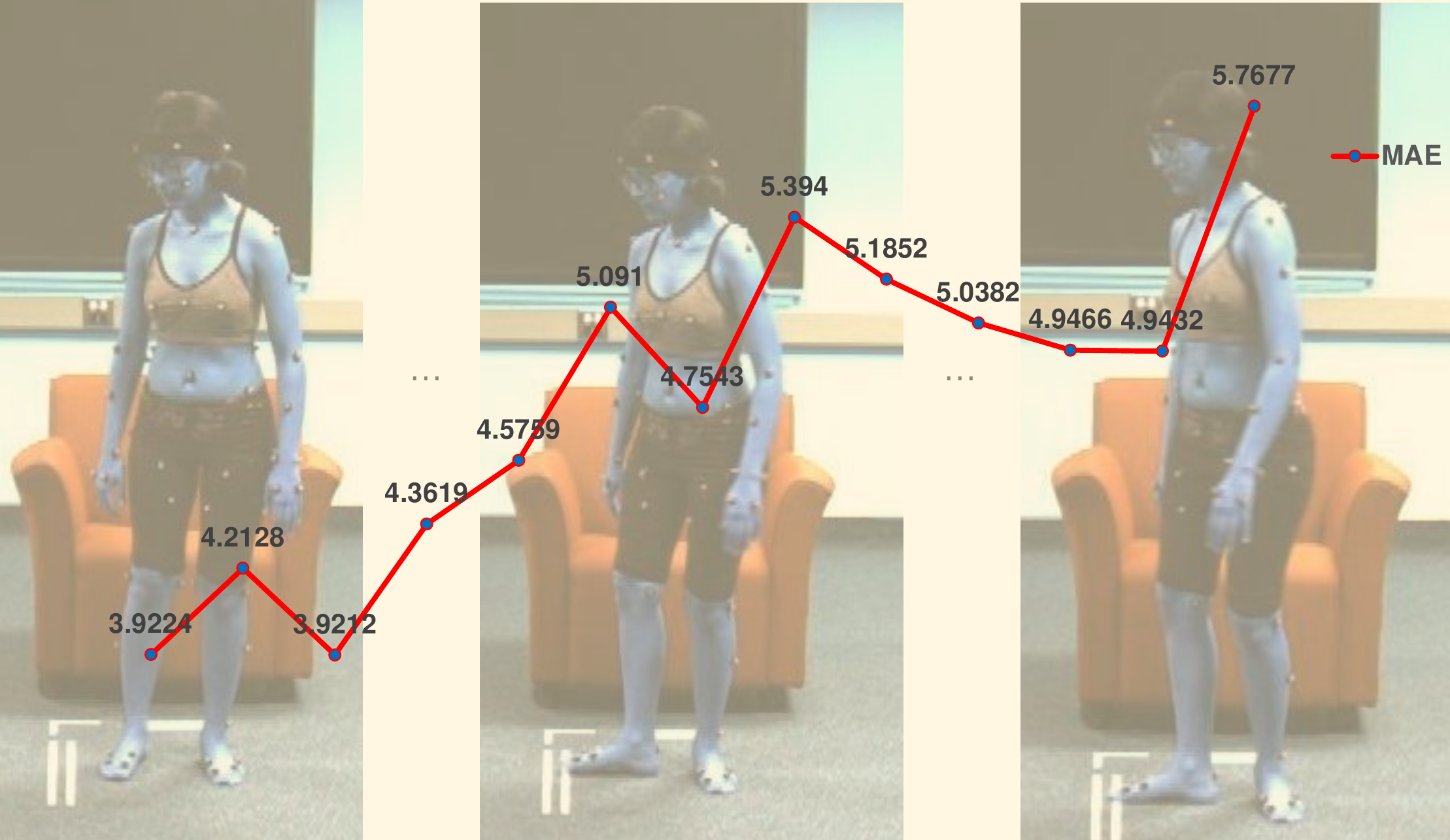}
	\caption{The effect of viewpoint on image-based method \cite{jin2022attention}}
	\label{view}
\end{figure*}

Moreover, we test sensitivity to the view angle of posture in images for the image-based model, whose results are shown in Figure \ref{view}. With the viewpoint changing from front-view to side-view, the prediction error of the image-based model becomes gradually larger. The results indicate that the image-based model is susceptible to viewpoint changes in posture, which increases the difficulty of image captures. The minimum and maximum MAEs of the image-based model are 3.9924 and 5.7677 respectively, while the MAE of our model is only 0.2600 for the same video sequence. The result validates that our model can obtain more effective and robust features from the entire video sequence, taking the information from many different angles into account by spatial-temporal encoder.

Among the testing results, we find that the image-based model always overestimates BMI for the people wearing black clothes, as shown in Figure \ref{black}. A possible reason is the weight is overestimated, considering the colour of clothes is the same as the processed background. Instead, our method can predict BMI accurately in this case. Therefore, the proposed approach is more applicable than the image-based model.
\begin{figure*}[H]
        \center
        \scriptsize
        \begin{tabular}{lll}
                \includegraphics[width=0.3\textwidth]{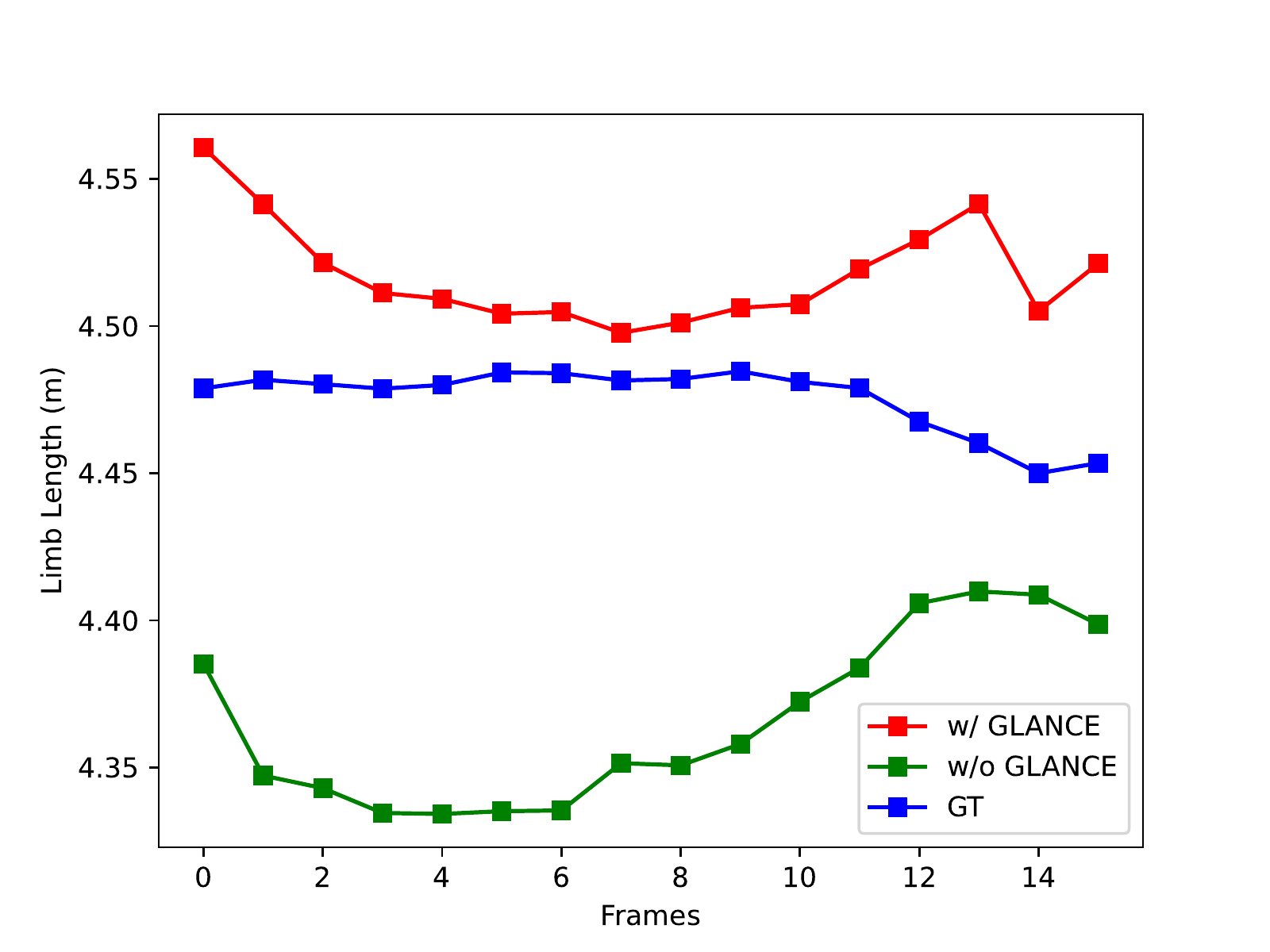} &    \includegraphics[width=0.3\textwidth]{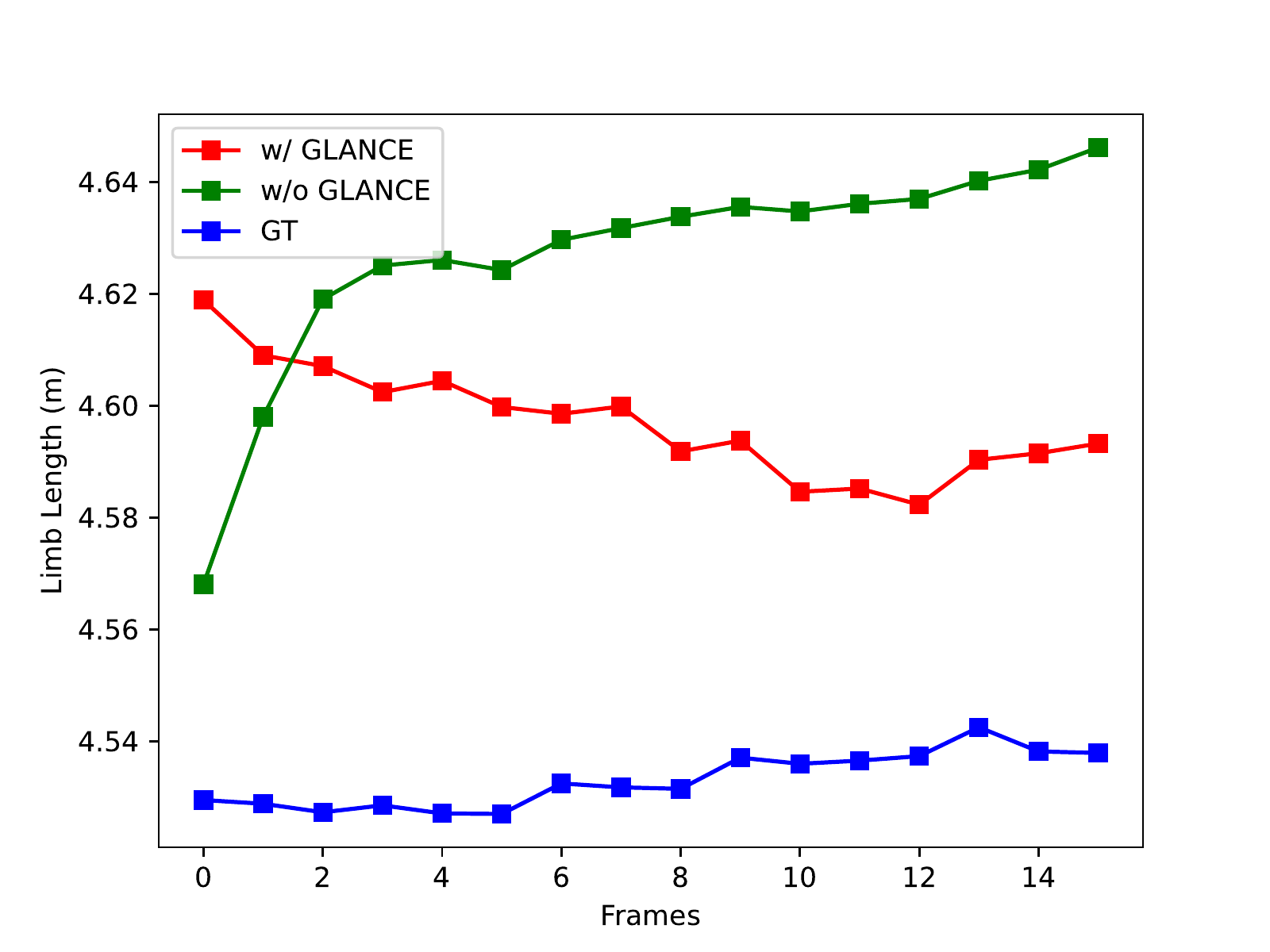} &
                \includegraphics[width=0.3\textwidth]{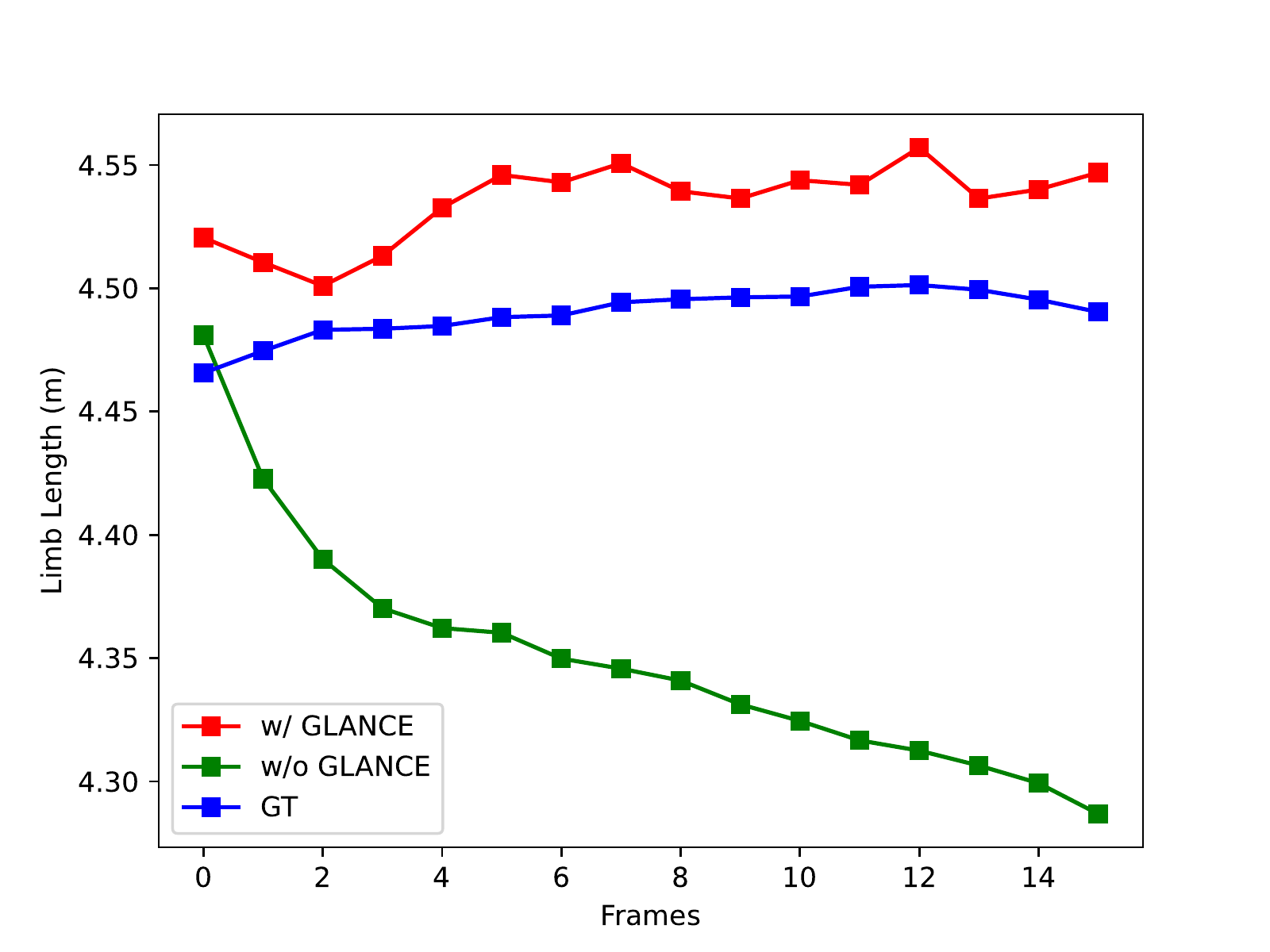}\\
                \\

        \end{tabular}
        \caption{The effectiveness of the GLANCE module for LimbLen variations in time.}
        \label{Limbvar}
        \vspace{-0.5em}
\end{figure*}
\begin{figure*}[H]
        \center
        \scriptsize
        \begin{tabular}{lll}
                \includegraphics[width=0.3\textwidth]{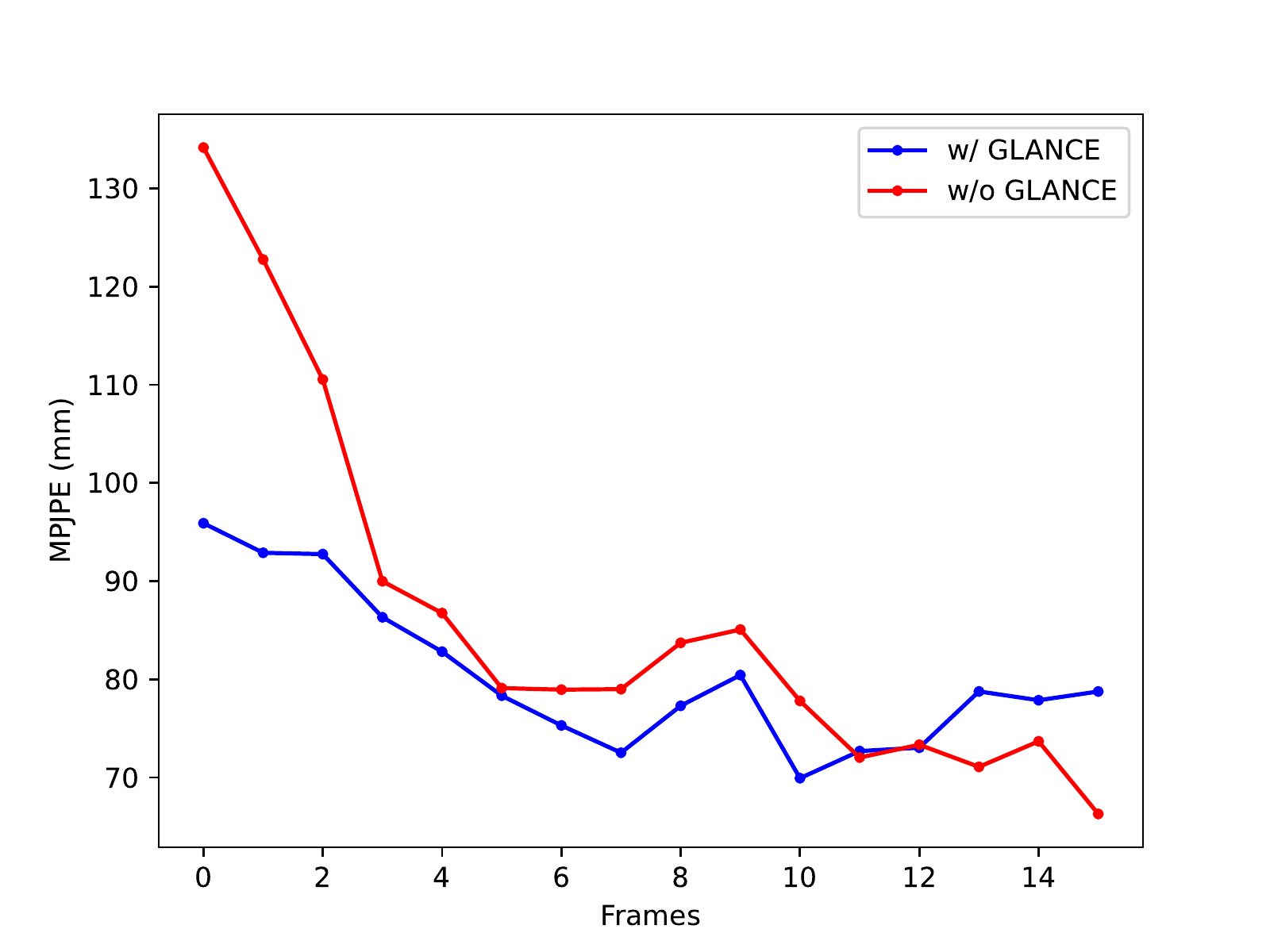} &    \includegraphics[width=0.3\textwidth]{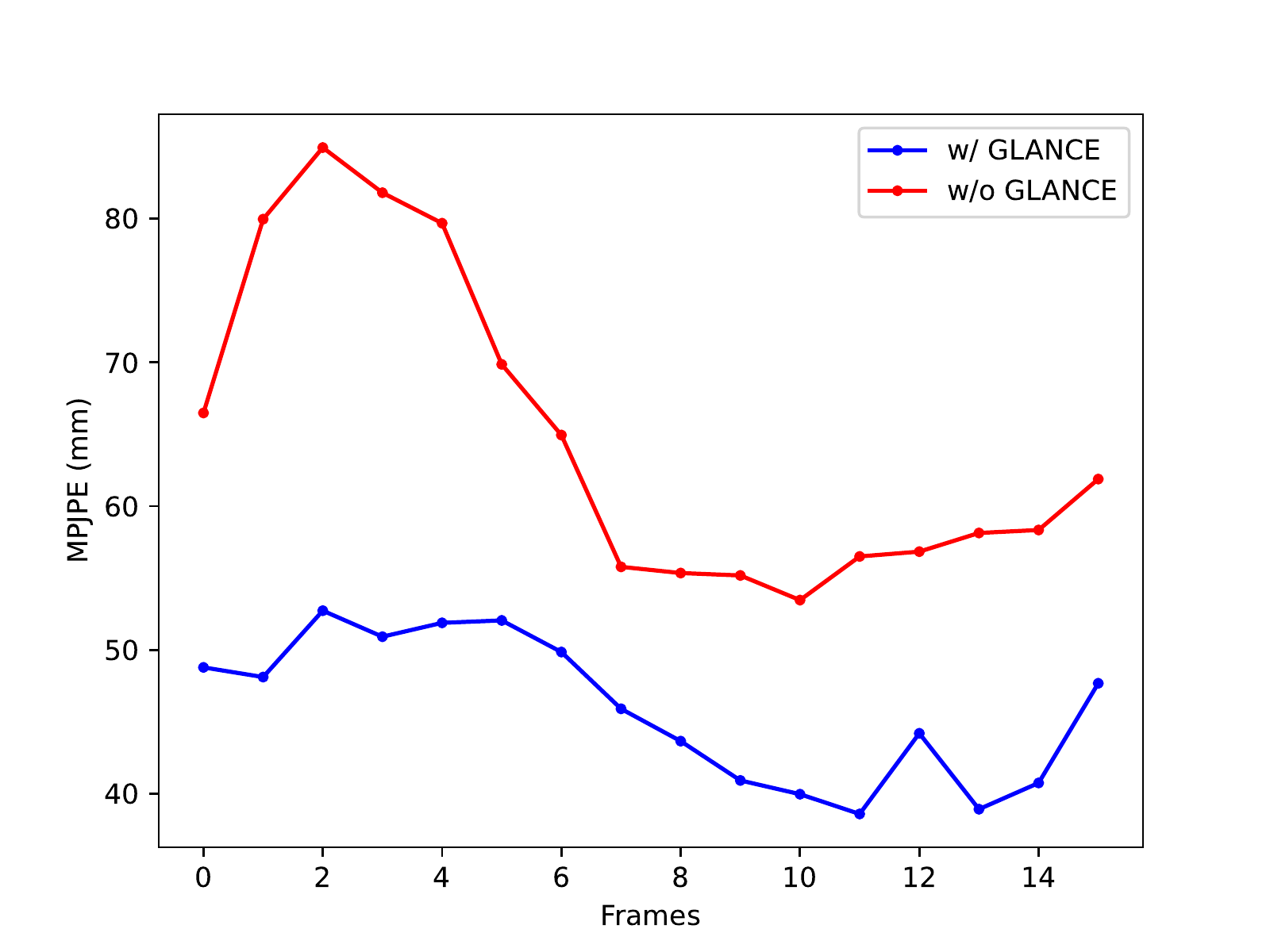} &
                \includegraphics[width=0.3\textwidth]{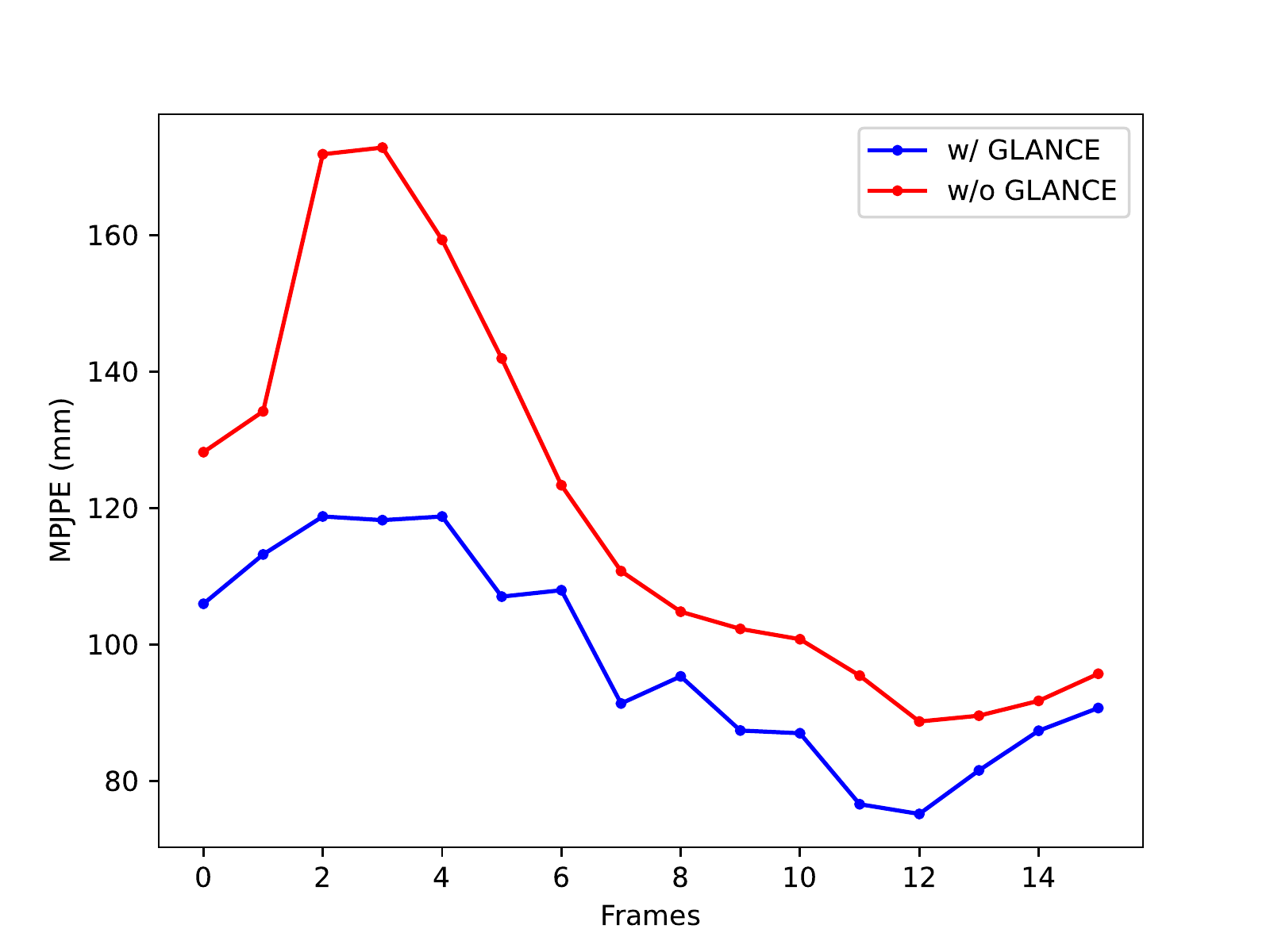}\\
                \\

        \end{tabular}
        \caption{The effectiveness of the GLANCE module for MPJPE variations in time.}
        \label{MPJPEvar}
        \vspace{-0.5em}
\end{figure*}

\subsection{Ablation Study}
In this section, we investigate the effectiveness of the proposed GLANCE module, which is divided into feature extractor and centrosymmetric fusion. The ablation study results of health indicator prediction are presented in Table \ref{ablation_health}. Compared with baseline model with only ResNet, which is actually VIBE\cite{kocabas_vibe_2020}, the performance is progressively enhanced by adding the feature extractor and the centrosymmetric fusion components. For the pose estimation task, the same conclusion is obtained from Table \ref{ablation_pose}. The gain indicates the effectiveness of the GLANCE module. Concretely, the feature extractor brings about global features, building upon the local feature map of ResNet. Then, the fusion component encourages local and global features to complement each other, further improving performance of the model.


Moreover, we also conduct some qualitative experiments to validate the effectiveness of the GLANCE module. Some cases of the variations of LimbLen over time are presented in Figure \ref{Limbvar}. The predictions of our model stay more closely to the targeted value than the one without GLANCE, which mainly attributes to extraction and integration of GLANCE module for multi-level features. This observation explains the excellent performance of our model on the health indicator prediction task since the LimbLen is directly related to health indicators. LimbLen error, bridging phase I and phase II, testifies the relationship between the two tasks by its resonant movements with the traditional pose estimation metric in Figure \ref{MPJPEvar}. The better the performance in pose estimation, the better it is for health indicator prediction. 

In Figure \ref{MPJPEvar}, the MPJPE of our model throughout a video is always less than the one without GLANCE. The reason for the GLANCE module to succeed is its ability to incorporate both global and local information and let each party learn from each other. In the case of occlusion of a joint due to viewpoint change, the global information of the relative position of a limb can help predict the location of the occluded joint using adjacent joint coordinates. The incorporation of global and local information helps the model cope with extreme circumstances and maintain competitive performance.

\section{Summary} \label{summary}
In this paper, we develop a paradigm for health indicator prediction from gait videos in small datasets. The spatial-temporal encoder for video feature extraction is first pre-trained on pose estimation, which is a related task to health indicator prediction. Then the pre-trained encoder is used for feature extraction in health indicator prediction. We also design the GLANCE module emphasizing global and local information extraction and fusion, whose effectiveness is validated on both tasks. A new pose estimation metric LimbLen Error is proposed to bridge the gap between tasks and show their connections. Experimental results show that our proposed paradigm is viable, and that the GLANCE module can improve performances in both tasks.

\section{Appendix}
\noindent\textbf{Declaration of Competing Interest:}We wish to confirm that there are no known conflicts of interest associated with this publication and there has been no significant financial support for this work that could have influenced its outcome. We confirm that the manuscript has been read and approved by all named authors and that there are no other persons who satisfied the criteria for authorship but are not listed. We further confirm that the order of authors listed in the manuscript has been approved by all of us. We confirm that we have given due consideration to the protection of intellectual property associated with this work and that there are no impediments to publication, including the timing of publication, with respect to intellectual property. In so doing we confirm that we have followed the regulations of our institutions concerning intellectual property. We understand that the Corresponding Author is the sole contact for the Editorial process (including Editorial Manager and direct communications with the office). He/she is responsible for communicating with the other authors about progress, submissions of revisions, and final approval of proofs. We confirm that we have provided a current, correct email address which is accessible by the Corresponding Author.

\noindent\textbf{Credit authorship contribution statement:}

\noindent\textbf{Ziqing Li: }conceptualization, software, data analysis, writing the original draft.
\textbf{Xuexin Yu: }conceptualization, editing, data analysis. 
\textbf{Xiaocong Lian: }Xiaocong Lian: conceptualization, editing, supervision.
\textbf{Yifeng Wang: }conceptualization
\textbf{Xiangyang Ji:} conceptualization, supervision, project administration.

\bibliographystyle{cas-model2-names}

\bibliography{cas-refs}

\begin{thebibliography}{37}
\expandafter\ifx\csname natexlab\endcsname\relax\def\natexlab#1{#1}\fi
\providecommand{\url}[1]{\texttt{#1}}
\providecommand{\href}[2]{#2}
\providecommand{\path}[1]{#1}
\providecommand{\DOIprefix}{doi:}
\providecommand{\ArXivprefix}{arXiv:}
\providecommand{\URLprefix}{URL: }
\providecommand{\Pubmedprefix}{pmid:}
\providecommand{\doi}[1]{\href{http://dx.doi.org/#1}{\path{#1}}}
\providecommand{\Pubmed}[1]{\href{pmid:#1}{\path{#1}}}
\providecommand{\bibinfo}[2]{#2}
\ifx\xfnm\relax \def\xfnm[#1]{\unskip,\space#1}\fi
\bibitem[{Andriluka et~al.(2018)Andriluka, Iqbal, Insafutdinov, Pishchulin,
  Milan, Gall and Schiele}]{andriluka2018posetrack}
\bibinfo{author}{Andriluka, M.}, \bibinfo{author}{Iqbal, U.},
  \bibinfo{author}{Insafutdinov, E.}, \bibinfo{author}{Pishchulin, L.},
  \bibinfo{author}{Milan, A.}, \bibinfo{author}{Gall, J.},
  \bibinfo{author}{Schiele, B.}, \bibinfo{year}{2018}.
\newblock \bibinfo{title}{Posetrack: A benchmark for human pose estimation and
  tracking}, in: \bibinfo{booktitle}{Proceedings of the IEEE conference on
  computer vision and pattern recognition}, pp. \bibinfo{pages}{5167--5176}.
\bibitem[{Artacho and Savakis(2020)}]{artacho2020unipose}
\bibinfo{author}{Artacho, B.}, \bibinfo{author}{Savakis, A.},
  \bibinfo{year}{2020}.
\newblock \bibinfo{title}{Unipose: Unified human pose estimation in single
  images and videos}, in: \bibinfo{booktitle}{Proceedings of the IEEE/CVF
  conference on computer vision and pattern recognition}, pp.
  \bibinfo{pages}{7035--7044}.
\bibitem[{Calvache et~al.(2020)Calvache, Bernal, Guarín, Aguía,
  Orjuela-Cañón and Perdomo}]{fall1}
\bibinfo{author}{Calvache, D.A.}, \bibinfo{author}{Bernal, H.A.},
  \bibinfo{author}{Guarín, J.F.}, \bibinfo{author}{Aguía, K.},
  \bibinfo{author}{Orjuela-Cañón, A.D.}, \bibinfo{author}{Perdomo, O.J.},
  \bibinfo{year}{2020}.
\newblock \bibinfo{title}{{Automatic estimation of pose and falls in videos
  using computer vision model}}, in: \bibinfo{editor}{Brieva, J.},
  \bibinfo{editor}{Lepore, N.}, \bibinfo{editor}{Linguraru, M.G.},
  \bibinfo{editor}{M.D., E.R.C.} (Eds.), \bibinfo{booktitle}{16th International
  Symposium on Medical Information Processing and Analysis},
  \bibinfo{organization}{International Society for Optics and Photonics}.
  \bibinfo{publisher}{SPIE}. pp. \bibinfo{pages}{281 -- 288}.
\newblock \URLprefix \url{https://doi.org/10.1117/12.2579615},
  \DOIprefix\doi{10.1117/12.2579615}.
\bibitem[{Cao et~al.(2017)Cao, Simon, Wei and Sheikh}]{openpose}
\bibinfo{author}{Cao, Z.}, \bibinfo{author}{Simon, T.}, \bibinfo{author}{Wei,
  S.E.}, \bibinfo{author}{Sheikh, Y.}, \bibinfo{year}{2017}.
\newblock \bibinfo{title}{Realtime multi-person 2d pose estimation using part
  affinity fields}, in: \bibinfo{booktitle}{Proceedings of the IEEE conference
  on computer vision and pattern recognition}, pp. \bibinfo{pages}{7291--7299}.
\bibitem[{Cho et~al.(2014)Cho, van Merrienboer, Bahdanau and Bengio}]{GRU}
\bibinfo{author}{Cho, K.}, \bibinfo{author}{van Merrienboer, B.},
  \bibinfo{author}{Bahdanau, D.}, \bibinfo{author}{Bengio, Y.},
  \bibinfo{year}{2014}.
\newblock \bibinfo{title}{On the properties of neural machine translation:
  Encoder-decoder approaches}.
\newblock \bibinfo{journal}{CoRR} \bibinfo{volume}{abs/1409.1259}.
\newblock \URLprefix \url{http://arxiv.org/abs/1409.1259},
  \href{http://arxiv.org/abs/1409.1259}{\tt arXiv:1409.1259}.
\bibitem[{Fu et~al.(2021)Fu, Liu, Yu, Chen and Wang}]{DWGAN}
\bibinfo{author}{Fu, M.}, \bibinfo{author}{Liu, H.}, \bibinfo{author}{Yu, Y.},
  \bibinfo{author}{Chen, J.}, \bibinfo{author}{Wang, K.}, \bibinfo{year}{2021}.
\newblock \bibinfo{title}{Dw-gan: A discrete wavelet transform gan for
  nonhomogeneous dehazing}, in: \bibinfo{booktitle}{Proceedings of the IEEE/CVF
  Conference on Computer Vision and Pattern Recognition}, pp.
  \bibinfo{pages}{203--212}.
\bibitem[{Ghorbani et~al.(2021)Ghorbani, Mahdaviani, Thaler, Kording, Cook,
  Blohm and Troje}]{ghorbani2021movi}
\bibinfo{author}{Ghorbani, S.}, \bibinfo{author}{Mahdaviani, K.},
  \bibinfo{author}{Thaler, A.}, \bibinfo{author}{Kording, K.},
  \bibinfo{author}{Cook, D.J.}, \bibinfo{author}{Blohm, G.},
  \bibinfo{author}{Troje, N.F.}, \bibinfo{year}{2021}.
\newblock \bibinfo{title}{Movi: A large multi-purpose human motion and video
  dataset}.
\newblock \bibinfo{journal}{Plos one} \bibinfo{volume}{16},
  \bibinfo{pages}{e0253157}.
\bibitem[{He et~al.(2015)He, Zhang, Ren and Sun}]{he2015delving}
\bibinfo{author}{He, K.}, \bibinfo{author}{Zhang, X.}, \bibinfo{author}{Ren,
  S.}, \bibinfo{author}{Sun, J.}, \bibinfo{year}{2015}.
\newblock \bibinfo{title}{Delving deep into rectifiers: Surpassing human-level
  performance on imagenet classification}, in: \bibinfo{booktitle}{Proceedings
  of the IEEE international conference on computer vision}, pp.
  \bibinfo{pages}{1026--1034}.
\bibitem[{Ionescu et~al.(2013)Ionescu, Papava, Olaru and
  Sminchisescu}]{human36}
\bibinfo{author}{Ionescu, C.}, \bibinfo{author}{Papava, D.},
  \bibinfo{author}{Olaru, V.}, \bibinfo{author}{Sminchisescu, C.},
  \bibinfo{year}{2013}.
\newblock \bibinfo{title}{Human3. 6m: Large scale datasets and predictive
  methods for 3d human sensing in natural environments}.
\newblock \bibinfo{journal}{IEEE transactions on pattern analysis and machine
  intelligence} \bibinfo{volume}{36}.
\bibitem[{Jin et~al.(2022)Jin, Huang, Xiong, Pang, Wang and
  Ding}]{jin2022attention}
\bibinfo{author}{Jin, Z.}, \bibinfo{author}{Huang, J.}, \bibinfo{author}{Xiong,
  A.}, \bibinfo{author}{Pang, Y.}, \bibinfo{author}{Wang, W.},
  \bibinfo{author}{Ding, B.}, \bibinfo{year}{2022}.
\newblock \bibinfo{title}{Attention guided deep features for accurate body mass
  index estimation}.
\newblock \bibinfo{journal}{Pattern Recognition Letters} \bibinfo{volume}{154},
  \bibinfo{pages}{22--28}.
\bibitem[{Kim et~al.(2018)Kim, Kook, Sun, Kang and Ko}]{fpn}
\bibinfo{author}{Kim, S.W.}, \bibinfo{author}{Kook, H.K.},
  \bibinfo{author}{Sun, J.Y.}, \bibinfo{author}{Kang, M.C.},
  \bibinfo{author}{Ko, S.J.}, \bibinfo{year}{2018}.
\newblock \bibinfo{title}{Parallel feature pyramid network for object
  detection}, in: \bibinfo{booktitle}{Proceedings of the European Conference on
  Computer Vision (ECCV)}, pp. \bibinfo{pages}{234--250}.
\bibitem[{Kingma and Ba(2014)}]{kingma2014adam}
\bibinfo{author}{Kingma, D.P.}, \bibinfo{author}{Ba, J.}, \bibinfo{year}{2014}.
\newblock \bibinfo{title}{Adam: A method for stochastic optimization}.
\newblock \bibinfo{journal}{arXiv preprint arXiv:1412.6980} .
\bibitem[{Klontz and Jain(2013)}]{bmiforid}
\bibinfo{author}{Klontz, J.C.}, \bibinfo{author}{Jain, A.K.},
  \bibinfo{year}{2013}.
\newblock \bibinfo{title}{A case study on unconstrained facial recognition
  using the boston marathon bombings suspects}.
\newblock \bibinfo{journal}{Michigan State University, Tech. Rep}
  \bibinfo{volume}{119}, \bibinfo{pages}{1}.
\bibitem[{Kocabas et~al.()Kocabas, Athanasiou and Black}]{kocabas_vibe_2020}
\bibinfo{author}{Kocabas, M.}, \bibinfo{author}{Athanasiou, N.},
  \bibinfo{author}{Black, M.J.}, .
\newblock \bibinfo{title}{{VIBE}: Video inference for human body pose and shape
  estimation}, in: \bibinfo{booktitle}{2020 {IEEE}/{CVF} Conference on Computer
  Vision and Pattern Recognition ({CVPR})}, \bibinfo{publisher}{{IEEE}}. pp.
  \bibinfo{pages}{5252--5262}.
\newblock \URLprefix \url{https://ieeexplore.ieee.org/document/9156519/},
  \DOIprefix\doi{10.1109/CVPR42600.2020.00530}.
\bibitem[{Kocabey et~al.(2017)Kocabey, Camurcu, Ofli, Aytar, Marin, Torralba
  and Weber}]{2017Face}
\bibinfo{author}{Kocabey, E.}, \bibinfo{author}{Camurcu, M.},
  \bibinfo{author}{Ofli, F.}, \bibinfo{author}{Aytar, Y.},
  \bibinfo{author}{Marin, J.}, \bibinfo{author}{Torralba, A.},
  \bibinfo{author}{Weber, I.}, \bibinfo{year}{2017}.
\newblock \bibinfo{title}{Face-to-bmi: Using computer vision to infer body mass
  index on social media}.
\bibitem[{Krizhevsky et~al.(2012)Krizhevsky, Sutskever and Hinton}]{Imagenet}
\bibinfo{author}{Krizhevsky, A.}, \bibinfo{author}{Sutskever, I.},
  \bibinfo{author}{Hinton, G.E.}, \bibinfo{year}{2012}.
\newblock \bibinfo{title}{Imagenet classification with deep convolutional
  neural networks}.
\newblock \bibinfo{journal}{Advances in neural information processing systems}
  \bibinfo{volume}{25}.
\bibitem[{Lin et~al.(2021)Lin, Wang and Liu}]{Metro}
\bibinfo{author}{Lin, K.}, \bibinfo{author}{Wang, L.}, \bibinfo{author}{Liu,
  Z.}, \bibinfo{year}{2021}.
\newblock \bibinfo{title}{End-to-end human pose and mesh reconstruction with
  transformers}, in: \bibinfo{booktitle}{2021 IEEE/CVF Conference on Computer
  Vision and Pattern Recognition (CVPR)}, pp. \bibinfo{pages}{1954--1963}.
\newblock \DOIprefix\doi{10.1109/CVPR46437.2021.00199}.
\bibitem[{Loper et~al.(2015)Loper, Mahmood, Romero, Pons-Moll and
  Black}]{2015SMPL}
\bibinfo{author}{Loper, M.}, \bibinfo{author}{Mahmood, N.},
  \bibinfo{author}{Romero, J.}, \bibinfo{author}{Pons-Moll, G.},
  \bibinfo{author}{Black, M.J.}, \bibinfo{year}{2015}.
\newblock \bibinfo{title}{Smpl: A skinned multi-person linear model}.
\newblock \bibinfo{journal}{Acm Transactions on Graphics} \bibinfo{volume}{34},
  \bibinfo{pages}{248}.
\bibitem[{Mehta et~al.(2018)Mehta, Sotnychenko, Mueller, Xu, Sridhar, Pons-Moll
  and Theobalt}]{Muco}
\bibinfo{author}{Mehta, D.}, \bibinfo{author}{Sotnychenko, O.},
  \bibinfo{author}{Mueller, F.}, \bibinfo{author}{Xu, W.},
  \bibinfo{author}{Sridhar, S.}, \bibinfo{author}{Pons-Moll, G.},
  \bibinfo{author}{Theobalt, C.}, \bibinfo{year}{2018}.
\newblock \bibinfo{title}{Single-shot multi-person 3d pose estimation from
  monocular rgb}, in: \bibinfo{booktitle}{2018 International Conference on 3D
  Vision (3DV)}, \bibinfo{organization}{IEEE}. pp. \bibinfo{pages}{120--130}.
\bibitem[{Meigs et~al.(2006)Meigs, Wilson, Fox, Vasan, Nathan, Sullivan and
  D’Agostino}]{meigs2006body}
\bibinfo{author}{Meigs, J.B.}, \bibinfo{author}{Wilson, P.W.},
  \bibinfo{author}{Fox, C.S.}, \bibinfo{author}{Vasan, R.S.},
  \bibinfo{author}{Nathan, D.M.}, \bibinfo{author}{Sullivan, L.M.},
  \bibinfo{author}{D’Agostino, R.B.}, \bibinfo{year}{2006}.
\newblock \bibinfo{title}{Body mass index, metabolic syndrome, and risk of type
  2 diabetes or cardiovascular disease}.
\newblock \bibinfo{journal}{The Journal of Clinical Endocrinology \&
  Metabolism} \bibinfo{volume}{91}, \bibinfo{pages}{2906--2912}.
\bibitem[{Moccia et~al.(2020)Moccia, Migliorelli, Carnielli and
  Frontoni}]{infant}
\bibinfo{author}{Moccia, S.}, \bibinfo{author}{Migliorelli, L.},
  \bibinfo{author}{Carnielli, V.}, \bibinfo{author}{Frontoni, E.},
  \bibinfo{year}{2020}.
\newblock \bibinfo{title}{Preterm infants’ pose estimation with
  spatio-temporal features}.
\newblock \bibinfo{journal}{IEEE Transactions on Biomedical Engineering}
  \bibinfo{volume}{67}, \bibinfo{pages}{2370--2380}.
\newblock \DOIprefix\doi{10.1109/TBME.2019.2961448}.
\bibitem[{Pascali et~al.(2016)Pascali, Giorgi, Bastiani, Buzzigoli, Henriquez,
  Matuszewski, Morales and Colantonio}]{PASCALI2016238}
\bibinfo{author}{Pascali, M.}, \bibinfo{author}{Giorgi, D.},
  \bibinfo{author}{Bastiani, L.}, \bibinfo{author}{Buzzigoli, E.},
  \bibinfo{author}{Henriquez, P.}, \bibinfo{author}{Matuszewski, B.},
  \bibinfo{author}{Morales, M.A.}, \bibinfo{author}{Colantonio, S.},
  \bibinfo{year}{2016}.
\newblock \bibinfo{title}{Face morphology: Can it tell us something about body
  weight and fat?}
\newblock \bibinfo{journal}{Computers in Biology and Medicine}
  \bibinfo{volume}{76}, \bibinfo{pages}{238--249}.
\newblock \URLprefix
  \url{https://www.sciencedirect.com/science/article/pii/S0010482516301445},
  \DOIprefix\doi{https://doi.org/10.1016/j.compbiomed.2016.06.006}.
\bibitem[{Renehan et~al.(2008)Renehan, Tyson, Egger, Heller and
  Zwahlen}]{renehan2008body}
\bibinfo{author}{Renehan, A.G.}, \bibinfo{author}{Tyson, M.},
  \bibinfo{author}{Egger, M.}, \bibinfo{author}{Heller, R.F.},
  \bibinfo{author}{Zwahlen, M.}, \bibinfo{year}{2008}.
\newblock \bibinfo{title}{Body-mass index and incidence of cancer: a systematic
  review and meta-analysis of prospective observational studies}.
\newblock \bibinfo{journal}{The lancet} \bibinfo{volume}{371},
  \bibinfo{pages}{569--578}.
\bibitem[{Rosso et~al.()Rosso, Agostini, Takeda, Tadano and
  Gastaldi}]{rossoInfluenceBMIGait2019}
\bibinfo{author}{Rosso, V.}, \bibinfo{author}{Agostini, V.},
  \bibinfo{author}{Takeda, R.}, \bibinfo{author}{Tadano, S.},
  \bibinfo{author}{Gastaldi, L.}, .
\newblock \bibinfo{title}{Influence of {{BMI}} on {{Gait Characteristics}} of
  {{Young Adults}}: {{3D Evaluation Using Inertial Sensors}}}
  \bibinfo{volume}{19}.
\newblock \DOIprefix\doi{10.3390/s19194221}.
\bibitem[{Shin et~al.()Shin, Chung, Kistler, Fitschen, Wilund and
  Sosnoff}]{shinEffectMuscleStrength2014}
\bibinfo{author}{Shin, S.}, \bibinfo{author}{Chung, H.R.},
  \bibinfo{author}{Kistler, B.M.}, \bibinfo{author}{Fitschen, P.J.},
  \bibinfo{author}{Wilund, K.R.}, \bibinfo{author}{Sosnoff, J.J.}, .
\newblock \bibinfo{title}{Effect of muscle strength on gait in hemodialysis
  patients with and without diabetes} \bibinfo{volume}{37},
  \bibinfo{pages}{29--33}.
\newblock \DOIprefix\doi{10.1097/MRR.0b013e3283643d76}.
\bibitem[{van~der Straaten et~al.()van~der Straaten, De~Baets, Jonkers and
  Timmermans}]{vanderstraatenMobileAssessmentLower2018}
\bibinfo{author}{van~der Straaten, R.}, \bibinfo{author}{De~Baets, L.},
  \bibinfo{author}{Jonkers, I.}, \bibinfo{author}{Timmermans, A.}, .
\newblock \bibinfo{title}{Mobile assessment of the lower limb kinematics in
  healthy persons and in persons with degenerative knee disorders: {{A}}
  systematic review} \bibinfo{volume}{59}, \bibinfo{pages}{229--241}.
\newblock \DOIprefix\doi{10.1016/j.gaitpost.2017.10.005}.
\bibitem[{Sun et~al.(2019)Sun, Xiao, Liu and Wang}]{hrnet}
\bibinfo{author}{Sun, K.}, \bibinfo{author}{Xiao, B.}, \bibinfo{author}{Liu,
  D.}, \bibinfo{author}{Wang, J.}, \bibinfo{year}{2019}.
\newblock \bibinfo{title}{Deep high-resolution representation learning for
  human pose estimation}, in: \bibinfo{booktitle}{Proceedings of the IEEE/CVF
  conference on computer vision and pattern recognition}, pp.
  \bibinfo{pages}{5693--5703}.
\bibitem[{Varol et~al.(2017)Varol, Romero, Martin, Mahmood, Black, Laptev and
  Schmid}]{surreal}
\bibinfo{author}{Varol, G.}, \bibinfo{author}{Romero, J.},
  \bibinfo{author}{Martin, X.}, \bibinfo{author}{Mahmood, N.},
  \bibinfo{author}{Black, M.J.}, \bibinfo{author}{Laptev, I.},
  \bibinfo{author}{Schmid, C.}, \bibinfo{year}{2017}.
\newblock \bibinfo{title}{Learning from synthetic humans}, in:
  \bibinfo{booktitle}{Proceedings of the IEEE conference on computer vision and
  pattern recognition}, pp. \bibinfo{pages}{109--117}.
\bibitem[{Velardo et~al.(2012)Velardo, Dugelay, Paleari and Ariano}]{imagebmi1}
\bibinfo{author}{Velardo, C.}, \bibinfo{author}{Dugelay, J.L.},
  \bibinfo{author}{Paleari, M.}, \bibinfo{author}{Ariano, P.},
  \bibinfo{year}{2012}.
\newblock \bibinfo{title}{Building the space scale or how to weigh a person
  with no gravity}, in: \bibinfo{booktitle}{2012 IEEE International Conference
  on Emerging Signal Processing Applications}, pp. \bibinfo{pages}{67--70}.
\newblock \DOIprefix\doi{10.1109/ESPA.2012.6152447}.
\bibitem[{Von~Marcard et~al.(2018)Von~Marcard, Henschel, Black, Rosenhahn and
  Pons-Moll}]{von20183dpw}
\bibinfo{author}{Von~Marcard, T.}, \bibinfo{author}{Henschel, R.},
  \bibinfo{author}{Black, M.J.}, \bibinfo{author}{Rosenhahn, B.},
  \bibinfo{author}{Pons-Moll, G.}, \bibinfo{year}{2018}.
\newblock \bibinfo{title}{Recovering accurate 3d human pose in the wild using
  imus and a moving camera}, in: \bibinfo{booktitle}{Proceedings of the
  European Conference on Computer Vision (ECCV)}, pp.
  \bibinfo{pages}{601--617}.
\bibitem[{Wang et~al.(2018)Wang, Wu, Herranz, van~de Weijer, Gonzalez-Garcia
  and Raducanu}]{Transfer_GAN}
\bibinfo{author}{Wang, Y.}, \bibinfo{author}{Wu, C.}, \bibinfo{author}{Herranz,
  L.}, \bibinfo{author}{van~de Weijer, J.}, \bibinfo{author}{Gonzalez-Garcia,
  A.}, \bibinfo{author}{Raducanu, B.}, \bibinfo{year}{2018}.
\newblock \bibinfo{title}{Transferring gans: generating images from limited
  data}, in: \bibinfo{booktitle}{Proceedings of the European Conference on
  Computer Vision (ECCV)}, pp. \bibinfo{pages}{218--234}.
\bibitem[{Wen and Guo(2013)}]{WEN2013392}
\bibinfo{author}{Wen, L.}, \bibinfo{author}{Guo, G.}, \bibinfo{year}{2013}.
\newblock \bibinfo{title}{A computational approach to body mass index
  prediction from face images}.
\newblock \bibinfo{journal}{Image and Vision Computing} \bibinfo{volume}{31},
  \bibinfo{pages}{392--400}.
\newblock \URLprefix
  \url{https://www.sciencedirect.com/science/article/pii/S0262885613000462},
  \DOIprefix\doi{https://doi.org/10.1016/j.imavis.2013.03.001}.
\bibitem[{Windham et~al.()Windham, Griswold, Wang, Kucharska-Newton, Demerath,
  Gabriel, Pompeii, Butler, Wagenknecht, Kritchevsky and
  Mosley}]{windhamImportanceMidtoLateLifeBody2017}
\bibinfo{author}{Windham, B.G.}, \bibinfo{author}{Griswold, M.E.},
  \bibinfo{author}{Wang, W.}, \bibinfo{author}{Kucharska-Newton, A.},
  \bibinfo{author}{Demerath, E.W.}, \bibinfo{author}{Gabriel, K.P.},
  \bibinfo{author}{Pompeii, L.A.}, \bibinfo{author}{Butler, K.},
  \bibinfo{author}{Wagenknecht, L.}, \bibinfo{author}{Kritchevsky, S.},
  \bibinfo{author}{Mosley, Jr., T.H.}, .
\newblock \bibinfo{title}{The {{Importance}} of {{Mid-to-Late-Life Body Mass
  Index Trajectories}} on {{Late-Life Gait Speed}}} \bibinfo{volume}{72},
  \bibinfo{pages}{1130--1136}.
\newblock \DOIprefix\doi{10.1093/gerona/glw200}.
\bibitem[{Wolk et~al.(2003)Wolk, Berger, Lennon, Brilakis and
  Somers}]{wolk2003body}
\bibinfo{author}{Wolk, R.}, \bibinfo{author}{Berger, P.},
  \bibinfo{author}{Lennon, R.J.}, \bibinfo{author}{Brilakis, E.S.},
  \bibinfo{author}{Somers, V.K.}, \bibinfo{year}{2003}.
\newblock \bibinfo{title}{Body mass index: a risk factor for unstable angina
  and myocardial infarction in patients with angiographically confirmed
  coronary artery disease}.
\newblock \bibinfo{journal}{Circulation} \bibinfo{volume}{108},
  \bibinfo{pages}{2206--2211}.
\bibitem[{Yousaf et~al.(2021)Yousaf, Hussein and Sultani}]{2021Estimation}
\bibinfo{author}{Yousaf, N.}, \bibinfo{author}{Hussein, S.},
  \bibinfo{author}{Sultani, W.}, \bibinfo{year}{2021}.
\newblock \bibinfo{title}{Estimation of bmi from facial images using semantic
  segmentation based region-aware pooling}.
\newblock \bibinfo{journal}{Computers in Biology and Medicine} .
\bibitem[{Zhong et~al.()Zhong, Rau and Yan}]{zhongApplicationSmartBracelet2018}
\bibinfo{author}{Zhong, R.}, \bibinfo{author}{Rau, P.L.P.},
  \bibinfo{author}{Yan, X.}, .
\newblock \bibinfo{title}{Application of smart bracelet to monitor
  frailty-related gait parameters of older {{Chinese}} adults: {{A}}
  preliminary study} \bibinfo{volume}{18}, \bibinfo{pages}{1366--1371}.
\newblock \URLprefix
  \url{https://onlinelibrary.wiley.com/doi/abs/10.1111/ggi.13492},
  \DOIprefix\doi{10.1111/ggi.13492}.
\bibitem[{Zhou et~al.(2019)Zhou, Barnes, Lu, Yang and Li}]{zhou2019penaction}
\bibinfo{author}{Zhou, Y.}, \bibinfo{author}{Barnes, C.}, \bibinfo{author}{Lu,
  J.}, \bibinfo{author}{Yang, J.}, \bibinfo{author}{Li, H.},
  \bibinfo{year}{2019}.
\newblock \bibinfo{title}{On the continuity of rotation representations in
  neural networks}, in: \bibinfo{booktitle}{Proceedings of the IEEE/CVF
  Conference on Computer Vision and Pattern Recognition}, pp.
  \bibinfo{pages}{5745--5753}.

\end{thebibliography}

\end{document}